\def\year{2021}\relax
\documentclass[letterpaper]{article} 
\usepackage{aaai21}  
\usepackage{times}  
\usepackage{helvet} 
\usepackage{courier}  
\usepackage[hyphens]{url}  
\usepackage{graphicx} 
\usepackage{natbib}  
\usepackage{caption} 
\usepackage{xcolor}
\renewcommand{\psi}{\textbackslash psi~}

\title{Enabling a Social Robot to Process Social Cues to Detect when to Help a User }
\author{
    Jason R. Wilson, Phyo Thuta Aung, Isabelle Boucher

}
\affiliations{
    \textsuperscript{\rm 1}Franklin \& Marshall College\\

    Lancaster, Pennsylvania  17603\\
    jrw@fandm.edu

}

\begin{document}

\maketitle

\begin{abstract}
It is important for socially assistive robots to be able to recognize when a user needs and wants help. 
Such robots need to be able to recognize human needs in a real-time manner so that they can provide timely assistance. We propose an architecture that uses social cues to determine when a robot should provide assistance.  Based on a multimodal fusion approach upon eye gaze and language modalities, our architecture is trained and evaluated on data collected in a robot-assisted Lego building task. By focusing on social cues, our architecture has minimal dependencies on the specifics of a given task, enabling it to be applied in many different contexts.  
Enabling a social robot to recognize a user's needs through social cues can help it to adapt to user behaviors and preferences, which in turn will lead to improved user experiences.
\end{abstract}

\section{Introduction}

For socially assistive robots, there is trade-off between helping too much and too little, helping too early or too late. Helping too little can make the robot seem ineffective or unreliable, leading to a diminished trust in the robot \cite{langer2019trust}.  Helping too much can be annoying, disrupt flow, and harm the user's autonomy \cite{Greczek2015, wilson2018supporting}  Simply waiting for when the user explicitly asks for help or makes a mistake may be insufficient and does not enable the agent to provide proactive or unsolicited assistance.  The challenge lies in recognizing when the user needs and wants assistance, often through implicit cues \cite{gorur2017toward}.  A user may employ a variety of nonverbal behaviors and other social signals that indicate that assistance may be needed.  For example, gaze patterns can be used to  when a user becomes disengaged \cite{sidner2005explorations, directionsrobot2014} or predict ingredients a user is going to select in a sandwich making task \cite{huang2015using}.  However, gaze patterns alone are not sufficient (e.g., a user may ask a question without shifting their gaze), and multiple modalities are required to holistically understand when a user needs assistance.

Combining social signals (i.e., gaze patterns and language) with task performance has been shown to be effective in recognizing when a user needs assistance \cite{reneau2020supporting}, but the previous approach processed all of the videos offline and leveraged manual annotations in the videos.  We seek to verify that similar inferences can be made in real-time and without human annotations.  Additionally, the prior work heavily relied on a task model, but we focus on an approach that has minimal dependencies on the specifics of the task.

In this paper, we present a real-time architecture that recognizes when a user needs assistance. Our approach automatically analyzes eye gaze and speech of a user and then fuses outputs from gaze and language models to detect whether the user needs help or not. 
Our main contribution in this paper is the development and validation of an architecture that uses social cues to detect when a social robot should help a user.
We proceed by first discussing related work in processing social cues and detecting when a user needs help.  We then provide a detailed description of the architecture.  Next, we describe the evaluation, including a data collection experiment in which users build a Lego structure.  We present and discuss the results of training and testing the models in the architecture and then provide a small demonstration of the architecture being applied in a cooking scenario. Finally, we conclude with limitations and future work

\section{Related Work}

There are many tasks for which socially assistive robots have been explored, from helping in post-stroke rehabilitation \cite{mataric2007socially, feingold2020social}, and managing medications for people with Parkinson's disease \cite{wilson2020challenges} to helping students learn language skills \cite{kory2019long}.  Often the focus may be developing functionality for user monitoring \cite{mataric2007socially,fasola2013socially}, retrieving objects \cite{fischinger2016hobbit}, providing reminders \cite{louie2014autonomous, fischinger2016hobbit}, and other skills for assisting a user.  
In developing these skills, it is often assumed that the robot has a mechanism to determine \textit{when} to help the user and thus the focus is on \textit{how} the robot would assist.  However, recognizing the appropriate times to for a robot to assist
can be critical to providing a quality user experience while also protecting the autonomy of the user \cite{wilson2020challenges}.

Some approaches to determining when to assist use planning-based approaches that are dependent on the specifics of a task \cite{wilson2018needbased, gorur2017toward}.  Similarly, a robot may monitor a user's actions to detect incorrect execution, such as improper exercise movements \cite{mataric2007socially, fasola2013socially}. 
Mistakes are useful indicators of with what a person needs help, but it is not necessarily a good indicator of when a person would like assistance.
Emotions and eye gaze, on the other hand, provide useful cues for determining  when a user would like assistance \cite{gorur2017toward, kurylo2019using}.  Additionally, these social cues remove some of the reasoning dependencies on the specifics of the task.  For example, social robots can interpret body posture and eye gaze to determine a user's level of engagement \cite{directionsrobot2014, sidner2005explorations}.

There are also many examples of non-robotic systems that provide assistance, and many of these also heavily rely on task specific cues.
In intelligent tutoring systems, they often rely on models of the task that allow them to detect a mistake \cite{ORourke2015} or a blocking state \cite{corbeil2020probabilistic}.
The Lumi{\`e}re project \cite{Horvitz1998Lumiere}, which inferred user's goals and intents through their actions, was the basis for the fateful ``Clippy'' assistance in Microsoft Office.  One of the major failure of Clippy was its inability to recognize when to assist, instead obnoxiously providing unwelcome guidance

Obviously, non-robotic systems like these operate in a different context than socially assistive robots, which can use its physical embodiment to create a social presence \cite{heerink2010relating}.  Just as humans use gesture, posture, gaze, facial expressions, and other forms of non-verbal communication with each other, 
robots and humans may also use the non-verbal features.
For example, when a robot points and looks at an object on a table, a user naturally looks at the object.  Similarly, users may use eye gaze to communicate with the robot. A sandwich building robot can interpret the user's gaze to determine which ingredient a user is going to select \cite{huang2015using}, and a medication assisting robot can detect when a user needs help completing the task \cite{kurylo2019using}.
Also, a directions robot and a robot exercise coach detect engagement using the position and orientation of a user's face \cite{directionsrobot2014, shao2019you}.  
Task-independent language can also be used by a robot.  For example, a tour-guide robot detects a users' emotions by processing the text of users' speech \cite{graterol2021emotion}.

\section{Architecture}
\label{sec:Architecture}

The goal of our architecture is to be able to do real-time recognition of when a user needs assistance by fusing together multiple modalities. 
The design builds off of the prior work in using eye gaze \cite{kurylo2019using} and multimodal fusion \cite{reneau2020supporting} for need recognition.  
While the previous work heavily relied on manual annotations that labeled the users' speech, gestures, and eye gaze, as well as the robot's speech, we move towards a more ecologically valid setup by automatically detecting the user's speech and gaze behavior by processing the audio and video streams in real-time.
We also seek to design an approach that is applicable to many tasks in which a social robot is assisting on a physical task that does not require the robot to physically intervene.
As such, our architecture does not include a model of the task, in contrast to the prior multimodal fusion work \cite{reneau2020supporting}.

\begin{figure}[h]
  \centering
  \includegraphics[width=\linewidth]{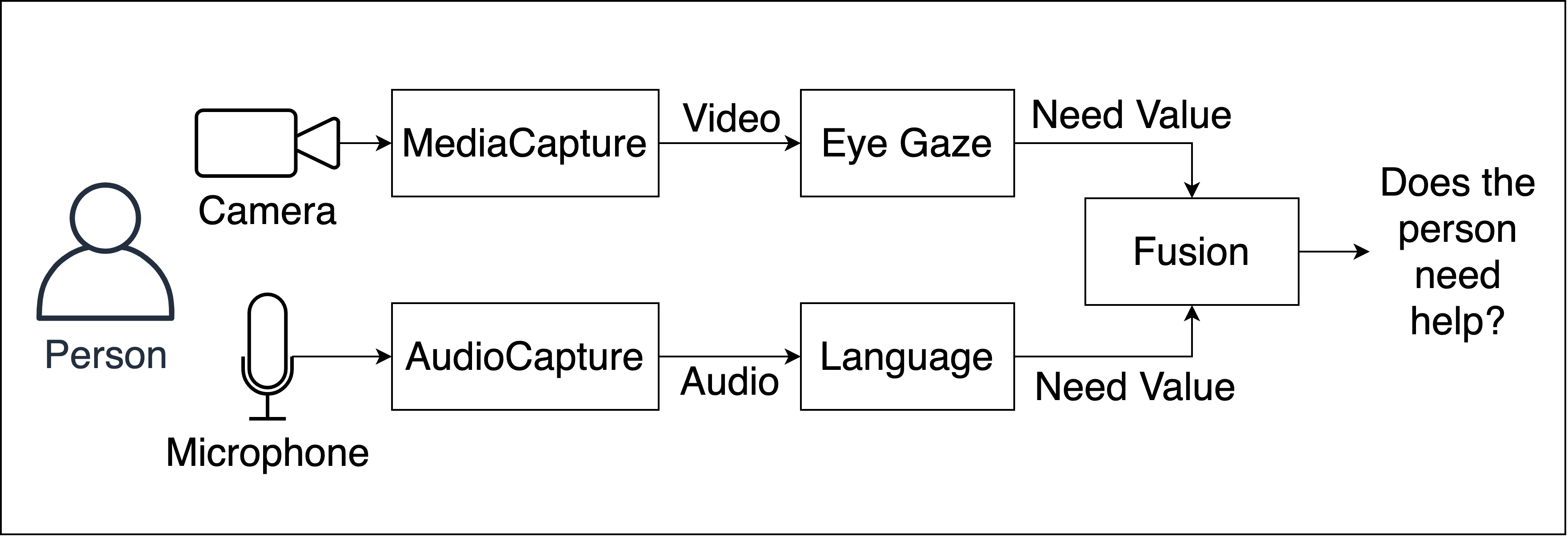}
  \caption{The architecture has two pipelines for audio and video processing, which are fused together to create the final output.}
  \label{fig:architecture} 
\end{figure}

As shown in Figure.~\ref{fig:architecture}, the system takes in video and audio inputs of a person through a camera and a microphone, and the data is then fed into the eye gaze component and the language component, respectively. The two components independently compute a need value, which is an estimate of how much assistance the user needs.  The need values are then used by the fusion component to determine whether the person needs help or not. To facilitate the processing of real-time data, we have built the system on Microsoft’s Platform for Situated Intelligence (\textbackslash psi) \cite{bohus2021platform}.

\subsection{Eye Gaze}
\label{sec:eyegaze}

\begin{figure}[h]
  \centering
  \includegraphics[width=\linewidth]{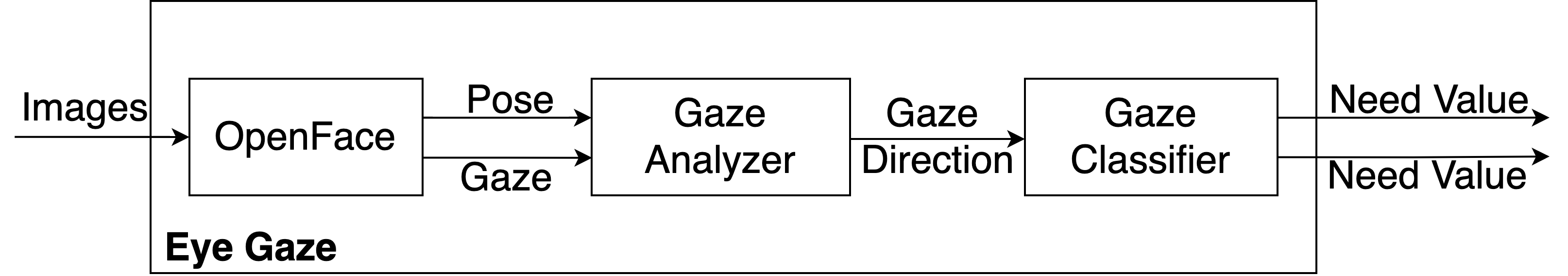}
  \caption{ The Eye Gaze component uses three subcomponents to produce two need values (one for each gaze pattern)}
  \label{fig:eyegaze}
\end{figure}

People often use eye gaze to help communicate their intentions.  There are many types of gaze patterns, such as \textit{mutual gaze}, which is used to communicate a desire to have the interlocutor attend to the same thing.  \textit{Confirmatory gaze} is similar, where a person confirms that the interlocutor is gazing upon the same thing.  These gaze patterns have also been shown to help predict when a user needs assistance \cite{kurylo2019using}.  Kurylo and Wilson analyzed four gaze patterns: mutual gaze, confirmatory gaze, gaze away (user looks away from robot and task), and goal reference (user looks at some object representing the goal of the task).  Mutual gaze and confirmatory gaze were found to be the best predictors of when the user needs assistance. The eye gaze component described here provides real-time prediction of when a user needs help by recognizing whether the user is exhibiting one of these two gaze patterns.

The eye gaze component (see Figure.~\ref{fig:eyegaze}) consists of three subcomponents: OpenFace, Gaze Analyzer, and Gaze Classifier.  We use OpenFace to extract vectors for eye gaze and head pose from images of the incoming video stream \cite{baltruvsaitis2016openface}. 
The quantitative values are mapped to a respective qualitative gaze direction by the Gaze Analyzer to describe whether the user is looking at the task, the robot, or elsewhere.  Qualitative gaze directions are Up, UpRight, Right, DownRight, Down, DownLeft, Left, UpLeft, and Center.  To focus on the problem of recognizing user need and not on identifying the locations of agents and tasks, the model makes a simple assumption about these locations. The robot is assumed to be directly in front of the user and the task is on the table in front of the user.  Center is then interpreted as looking at the robot, and down directions are interpreted as looking at the task. 

The qualitative eye gaze direction is passed onto the Gaze Classifier subcomponent, which contains two models, one to recognize a mutual gaze pattern and one for confirmatory gaze. Mutual gaze represents the case when the user initiates a mutual gaze with the agent (the robot) by redirecting its gaze to the agent. The confirmatory gaze represents the case when the subject is looking back and forth between the task and the agent. This case consists of two possible situations: (1) the subject briefly looks at the task and then briefly looks at the agent, and (2) the subject briefly looks at the agent and then briefly looks at the task. A brief glance is considered to be less than 2.5 seconds, as defined through experimentation in the prior work \cite{kurylo2019using}. 

These models differ in two ways from the prior work.  First, our current models are not sensitive to whether the robot is currently speaking.  Previously, the fact that the robot is speaking would prevent the models from inferring that the user needs help because the user is believed to be looking at the robot to attend to what it is saying, not because the user needs help.  The robot's speech is not currently integrated into the architecture of the system.  One effect of this simplification is that the model is now better described as not detecting that the user needs assistance but that the user is exhibiting one of the gaze patterns.  

The other modification is that the output of each of the models is a continuous value 0.0 to 1.0, where greater than 0.5 can be interpreted as exhibiting that gaze behavior.  To get a continuous value, we use the duration of the current gaze direction, and compare it to a 2.5 second threshold.
The output of a given model is then $min(1, d/2.5)$, where $d$ is the duration of the current gaze direction.
As a result, if the current direction of the eye gaze is more than 1.25 seconds, then the corresponding model (either mutual gaze or confirmatory gaze) will output a value of at least 0.5, meaning that the user is believed to be performing the given gaze pattern.


\subsection{Language}
\label{subsec:Language}

\begin{figure}[h]
  \centering
  \includegraphics[width=\linewidth]{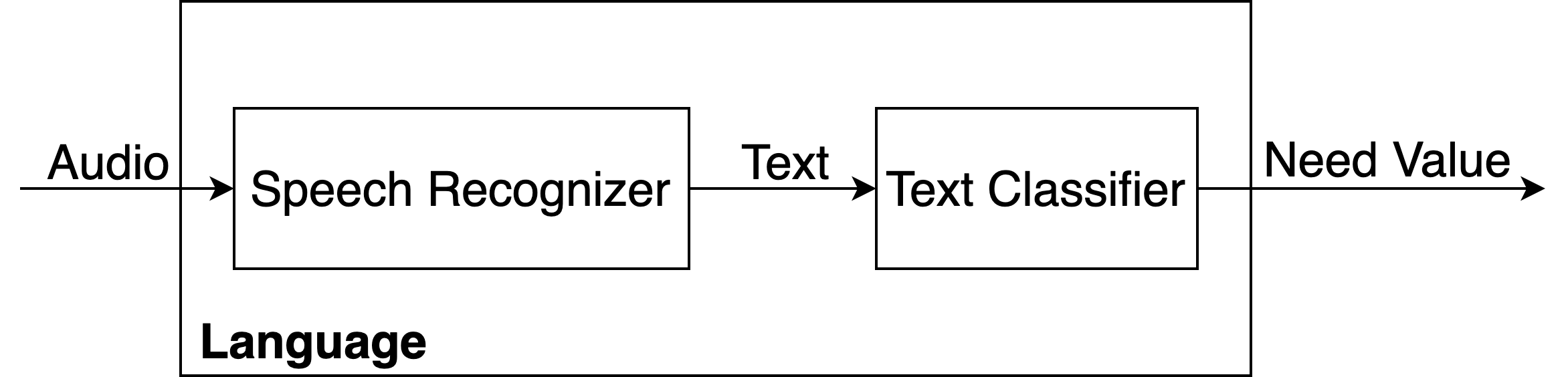}
  \caption{The Language component has two subcomponents to process the audio input and produce a need value.}
  \label{fig:language} 
\end{figure}

The user's language can often be used identify when the user needs help.  The user may make explicit requests for assistance (e.g., ``Please help me''), and the user may also make other references that can indicate when the user needs help (e.g., ``Hm, I'm not sure.'').  Prior work has used simple keyword spotting to recognize likely indicators \cite{reneau2020supporting}, but this can be too task-specific and not robust enough.  Instead, our Language component uses a learned model to classify utterances based on whether the user needs help.

The Language component decomposes into two parts: Speech Recognizer and Text Classifier (see  Figure.~\ref{fig:language}).  The Speech Recognizer processes the raw audio input to produce the text of what the user says.  The Speech Recognizer is psiDeepSpeech, a \psi component for the Deep Speech\footnote{https://github.com/mozilla/DeepSpeech} (v0.9.3) speech-to-text engine.  The text is passed to the Text Classifier, which contains a Naïve Bayes model. We chose Naïve Bayes as a model because of its data efficiency, as there is currently a limited amount of training data currently available. 
The model includes two features beyond the individual words in the text.  We aggregate all question words (i.e., what, who, which, where, when, how, why)  into an additional feature.  We also have a feature for negations, which includes no, not, none, nothing, isn't, aren't, don't, won't, wasn't, weren't, wouldn't, shouldn't, couldn't, can't.  We added these features because questions and negations are expected to be indicative of a person needing assistance. 

\subsection{Fusion}
\label{subsec:Fusion}
The Fusion component makes the final determination of whether a person needs help based on the outputs of the Eye Gaze and Language components. 
Each model on its own is insufficient.  The Eye Gaze component cannot capture moments when the user is verbally requesting help, and the Language component cannot recognize the unspoken cues that indicate a user needs assistance.  Additionally, recent behavior also can help interpret the outputs of these components.  For example, a confirmatory gaze pattern shortly followed by an ``okay'' utterance could mean the user is attempting to ask if what the user is looking at is correct.  Our approach to combining the outputs of each modality while also taking into account temporal dependencies is to use a decision-level fusion with a sliding window, based on results from prior work on mulitmodal fusion to detect need \cite{reneau2020supporting}.  


The outputs of the gaze and language models represent an inference for a single moment in time.  The language model does not incorporate any prior utterances, and the gaze models have only a limited sense of the recent gaze directions.  However, previous events should inform how future events are interpreted.  To capture these temporal dependencies, we use a sliding window, which concatenates a number (equal to the window size) of the  outputs of the specialized models together into a single feature vector as input to the fusion model. 
Sliding windows are commonly used to provide more temporal context.  For example, 
the method has been used for action recognition for industrial applications \cite{akkaladevi2015action} and gesture-based recognition in warehouse robots \cite{neto2019gesture}.
Our use of the sliding window method concatenates together some number (the window size) of prior outputs from the gaze and language models.  Our current window size is 20, which is less than the optimal size of 47 reported in \cite{reneau2020supporting}.  However, a window size of 20 in the prior work did not perform significantly worse while providing benefits of speed improvements.

While implementing a sliding window can be relatively when using the built-in functionality of \psi, a challenge is that the language model produces outputs far less frequently than the gaze models.  Since we do not want to wait for the next language model output but also want our feature vector to represent all of the input models, we apply an interpolation function to the output of the language model, which allows us to get a language model output at a frequency more comparable to the gaze models.



The new feature vector is then fed into fusion model, for which we use a random forest classification, which has previously outperformed decision trees, naïve Bayes, support vector machine, and logistic regression in a similar problem. \cite{reneau2020supporting}. The random forest model predicts a final output need value, which determines whether a person needs assistance.

\section{Evaluation}

To evaluate how well the architecture is able to recognize when a person is exhibiting social cues indicating that a user needs assistance, we designed an experiment to collect data that is used to train and evaluate the system.
The following criteria was used in designing the experiment:
\begin{itemize}
    \item The task performed by the user (and assisted by the user) must require assistance that can be purely social or cognitive.  The task cannot require or utilize physical intervention from the robot
    \item The task needs to have some physical presence.  It cannot be purely mental or virtual (relying primarily on an electronic interface).  The physical task encourages more user engagement, promotes more changes in eye gaze direction, and has more opportunities to leverage the robot's physical presence even though it would not provide any physical support in the task.
    \item The task needs to be challenging enough that participants would generally need some assistance.  
\end{itemize}

For the experiment, each participant assembles a Pterodactyl with Lego pieces.  Participants given a set of pieces, some of which are already partially assembled (see Figure~\ref{fig:exp_setup}).  They are not provided with assembly instructions and instead are given a black-and-white photo of the final structure.  The intent is to give participants many paths to the correct assembly and not provide too much guidance (i.e., step-by-step instructions) but also add some challenge (i.e., not using a color picture).  Additionally, to increase the likelihood that each participant will require at least some assistance, one piece was hidden in a pouch on the table.

\subsection{Data Collection}

\begin{figure}[t]
    \centering
    \includegraphics[width=\columnwidth]{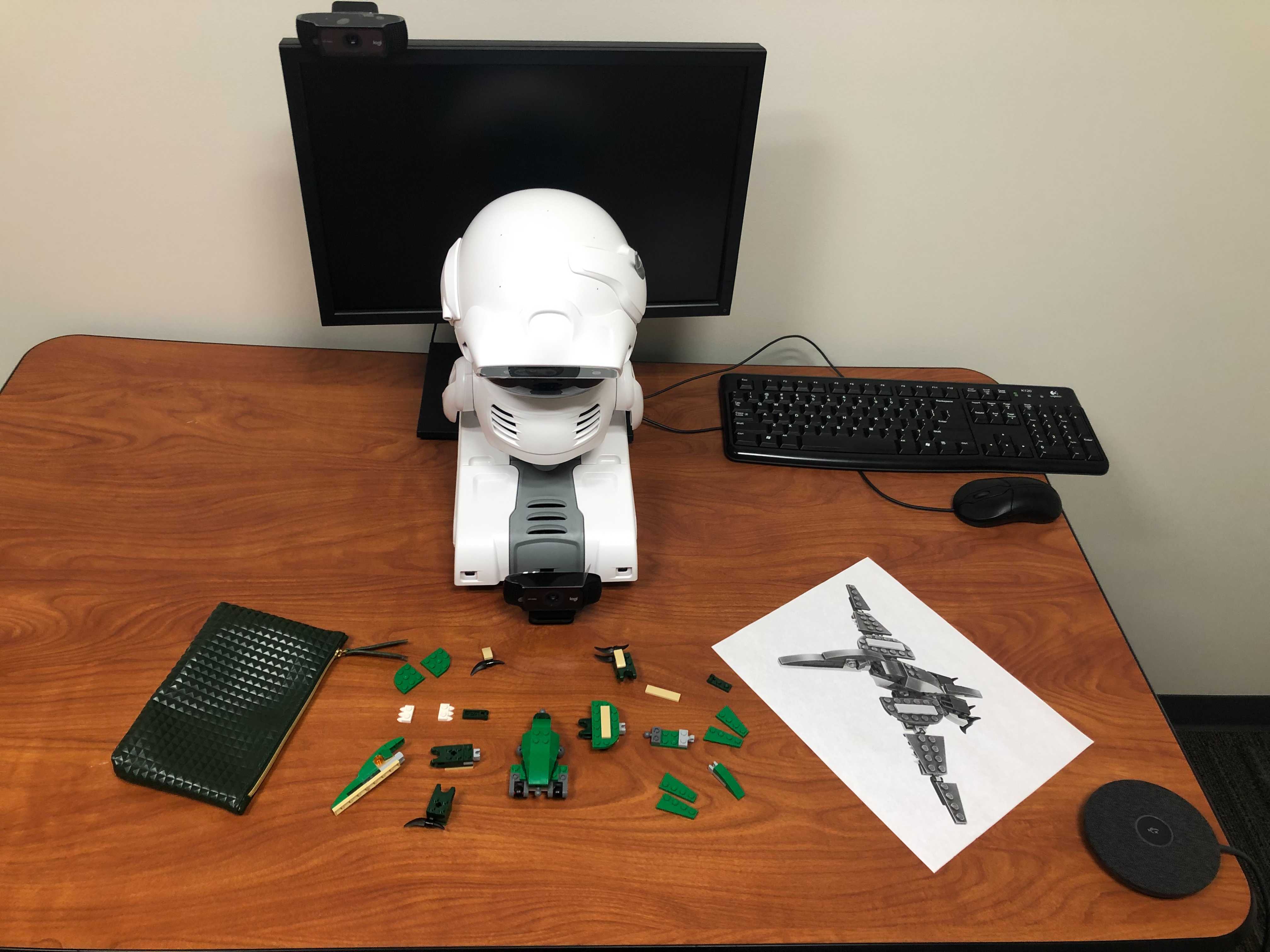}
    \caption{The Lego pieces, some of which were partially assembled, was on the table directly between the robot and the participant.  The photo on the participant's right was the structure they were asked to build.  The pouch on the left contained a missing piece.}
    \label{fig:exp_setup}
\end{figure}

Twenty-one participants interacted with a socially assistive robot as they completed the Lego building task.  See Figure~\ref{fig:exp_setup} for the physical layout. The participants were seated in front of a robot placed on the table in front of them. Two webcams, one camera mounted on top of a monitor behind the robot and another placed in front of the robot angled upwards, were used.  The microphone on one webcam provided the audio input. Each participant was provided with a partially disassembled Lego structure and a printed black-and-white image of the intended structure.

The experiment was conducted with a Wizard-Of-Oz setup with a research assistant observing the interaction from a neighboring room and remotely controlling the robot to assist the participants throughout the task. The robot provided three main types of responses: verbal encouragement, indirect feedback, and direct feedback \cite{RogersHolm1994}. Verbal encouragement included phrases such as ``everything looks good so far'' and ``yes, you are on the right track.'' Both indirect and direct feedback referenced specific components of the task but varied on the intensity of the assistance. An example of indirect feedback includes a phrase such as ``you are close but there is something that is not correct.'' An example of direct feedback includes ``there is a problem with the wing'' or feedback specifying the Lego pieces that make up a part of the structure after the participant requests more in depth assistance.  The robot also altered its gaze as needed -- looking at the user, the Lego pieces, and the picture.  

The interaction began with an orientation phase which involved the robot requesting the participant to point to objects on the table, such as a travel mug and a pouch, and the robot turning its head in the specified direction and providing feedback. Once the participant completed the orientation phase the robot instructed the participant to begin building the structure shown in the image on the table.

The experiment was designed to optimize different gaze directions as well as encourage the participant to speak to the agent. This involved utilizing a physical agent as well as a physical task, Lego building, and providing a physically printed image. The orientation phase was not analyzed. The pouch used in the orientation phase enclosed one of the essential pieces of the structure. The intention of this component was to increase the chances that the participant would speak to the robot at some point during the experiment. 

After completing the Lego task, each participant was asked to complete a questionnaire to reflect on their experience interacting with the robot and their confidence in the Lego task.  The questionnaire included 6 items with a 4-point Likert scale (strongly/somewhat disagree/agree), one multiple choice item, one item where they ranked options, and one free response item.  The content of the questionnaire is as follows:

\begin{enumerate}
    \item I wanted to engage with the robot throughout the task.
    \item I found the interaction with the robot to be engaging.
    \item I trusted the robot to assist me during the task.
    \item I was confident in my capabilities for completing the task.
    \item Other than asking about the missing piece, were there any times when you needed the robot’s assistance? Please specify these instances.
    \item I found the robot to be a helpful resource when attempting the task.
    \item The robot provided [too much, the correct amount, too little] assistance.
    \item The robot enabled me to complete the task to the best of my ability.
    \item Please rank the following resources that you relied on to complete the task [Previous Lego building experience, Picture of final Lego structure, Robot, Trial and error]
\end{enumerate}

\subsection{Labeling}

All audio/video data was labeled using five labels, each describing a different level of need that the user required at a specified time interval. The levels of need were defined based on social cues such as eye gaze and speech exhibited by the user throughout the experiment. We distinguished a difference between the participants' status in the task and their awareness and mental state while completing the task.  This is because a person may be working on a task, though not making progress, but be actively working towards a solution.  For this state, we consider the user to be a in a flow state \cite{csikszentmihalyi1990flow} and does not need help from the robot.  The level of need is dependent on the presence of  flow and the resources they seek to utilize during a specified time interval. When the user exits the flow states, there are increasing levels of need.

\begin{figure}[h]
    \centering
    \includegraphics[width=\columnwidth]{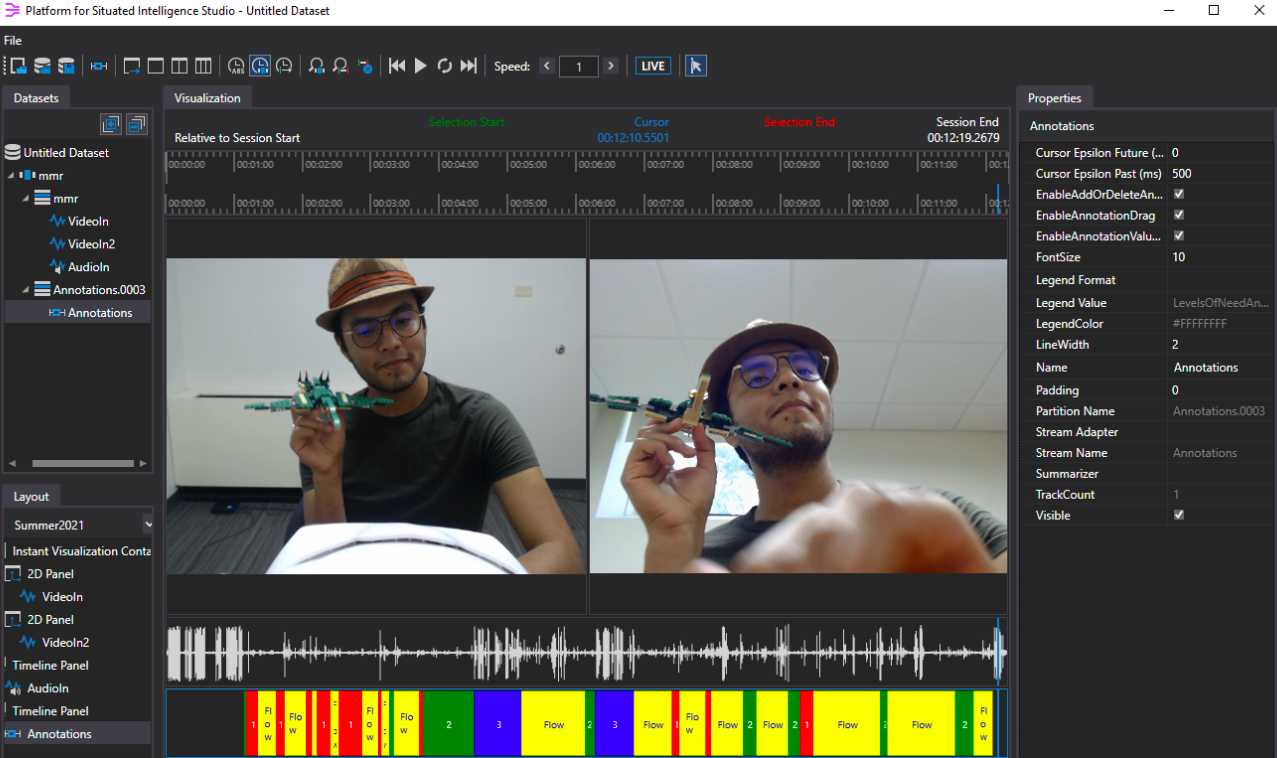}
    \caption{Every second of the data for each participant was labeled.  The annotator had access to the two video streams and the audio that was recorded.}
    \label{fig:annotations}
\end{figure}

For each video, all time after the orientation phases was labeled with one of five need levels (see Figure~\ref{fig:annotations} for an example).  Levels of help 1-3 are intended to mirror the first three levels of assistance \cite{RogersHolm1994}, based on a theory that when the level of assistance matches the level of need, human autonomy can be maximized/supported \cite{wilson2018dignity}.
The following criteria was used when labeling the data.
\begin{itemize}
    \item \textbf{Flow} represents when the participant is progressing through that task at a steady pace and appears to show no signs of distress, signifying a steady mental state. 
    \item \textbf{Level 0} is used when the participant is not progressing through the task but does not show any social cues signifying that they are in need of assistance. Level 0 is typically used to represent the transition from Flow to level 1. 
    \item \textbf{Level 1} represents a confirmatory stage, when a participant exhibits social cues such as looking around the table or back and forth from the picture to the structure, or pausing the progress in the task. 
    \item \textbf{Level 2} represents when the participant is aware that the Flow is broken and they resolve the issue using means other than themselves and the robot such as using the picture, another piece. Oftentimes the participant gaze switches back and forth from the structure and the image and they orient their body towards the images rather than facing the robot.  
    \item  \textbf{Level 3} is similar to level 2 and is when the participant is aware that flow is broken.  They also deem all other sources are not sufficient and therefore they verbally seek assistance from the robot. 
\end{itemize}

\subsection{Training the System}
\label{subsec:Training the System}

We provide details of the process for training the models to demonstrate how the architecture is leveraged in both the training and execution. 
This is intended to show the relative ease with which the architecture could be extended to integrate additional models. 
One of the greatest advantages of basing the architecture on \psi is the automatic synchronization of data streams. It is vital that the streams being fused together be correctly synchronized and that the synchronization scheme is consistent in collecting the training data and executing the live system. 

To evaluate the architecture, we train the models in two stages. In the first stage, we train the language model.  The gaze models are rule-based and do not require training.  In the second stage, we use the output of the trained language model in collecting the data to train the fusion model.  It is important that, when training the fusion model on the outputs of the first layer of models, the outputs are synchronized based on the times of when the data originated (not when the data is outputted by the most recent model).  To maintain the synchronization based on originating times, we leverage the inherent capabilities of \psi.  All of training and intermediate data is managed by \psi, which is designed to then handle all of the data synchronization. 
Training of each of the models, on the other hand, is assumed to be done with specialized tools outside of \psi (e.g., scikit-learn).
Thus, the training process includes steps to export the data to an appropriate format to be processed by the machine learning tool.

First, all data is collected into a \psi data store (DS0).  Then there is the following two stages:
\begin{enumerate}
    \item Train specialized models
    \begin{enumerate}
        \item Export data from \psi data store (DS0) to format for training the language model.
        \item Train model on exported data and integrate resulting model into \psi component.
        \item Generate new \psi data store (DS1) with component outputs by running architecture on original data store.
    \end{enumerate}
    \item Train fusion model
    \begin{enumerate}
        \item Export data from data store (DS1) to format for training the model.  This step includes applying the sliding window to create feature vectors containing past model outputs.
        \item Train fusion model on exported data and integrate resulting model into \textbackslash psi component.
    \end{enumerate}
\end{enumerate}

When training the language model to be run with the full architecture, we use all of the available data, which is the text outputted from the speech recognizer.  However, we also want to assess how the language models works in isolation, and we do a 10-fold cross validation to assess this.  Similarly, in evaluating the performance of the fusion model, we also conduct a 10-fold cross validation.

The process described above can be easily extended to include additional models.  The process will remain a two stage process, where the first stage trains the models that precede the fusion model, and the second stage trains the fusion model.

\subsection{Results}

The experiment resulted in a total of 146.2 minutes of labeled video and audio, with 74.3 minutes of the data being labeled as when the participant needs some level of help. The average length of time it took the participants to complete the task was 7 minutes with a standard deviation of 4.2 minutes. Out of the 21 participants, 4 participants completed the task without speaking to the robot at any time throughout the experiment.

\begin{figure}
    \centering
    \includegraphics[width=85mm]{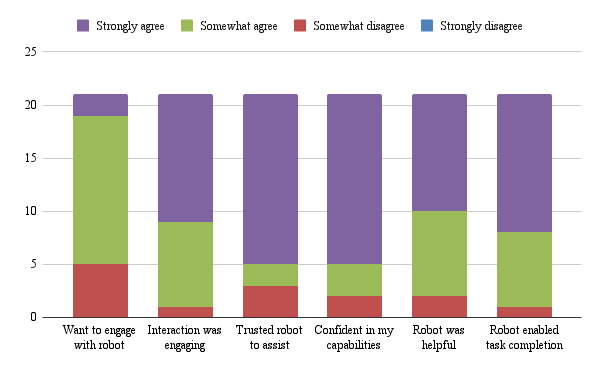}\\
    \includegraphics[width=22mm]{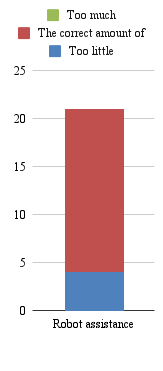}
    \includegraphics[width=60mm]{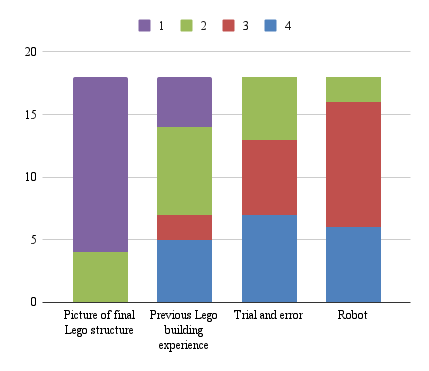}
    \caption{The top chart shows participants agreements with statements about the robot.  The bottom left chart shows participants judgments about how much help the robot provided.  The bottom right is the ranking of the resources that participants used.  Three participants chose not to rank the resources.}
    \label{fig:survey_results}
\end{figure}

The experiment was successful in engaging with the robot to get assistance on the task (see Fig.~\ref{fig:survey_results}).  In the questionnaire, participants agreed (somewhat or strongly) with statements about the robot being engaging (Q5, 95\% agree), trustworthy (Q6, 86\% agree), and helpful (Q8, 90\% agree).

As a measurement of the validity of the labels we compare the average level of help in each video to whether the participant indicated whether there was any item other than the missing piece for which they sought the robot's help (item 9 in the questionnaire).  To get the average help in each video, we calculated a weighted sum (level of need $\cdot$ seconds of label) of the labels, using a value of -1 for the Flow label, and divided by the total seconds.  Average help ranged from -0.49 to 2.24 ($\mu = 0.32, \sigma=0.55$).  We found that the average help correlated with whether the participant indicated they needed help on more than the missing piece ($r = .47, p = .03$).  This shows that the labels correspond with participants self-reports, giving some evidence for the validity of the labels.


To evaluate the architecture, we use the collected data to train the language and fusion models and then assess the performance of the output of the fusion model.
We report (in Table~\ref{tab:results} the performance of the trained language and fusion models, as well as the rule-based eye gazes model.  For the trained models, we report the results of a 10-fold cross validation.  Since the gaze models do not require training data, we report results using all of the recorded data.
All analysis is based on the models performing as binary classifiers: need the robot's help or not.  We interpret need levels 2 and 3 as needing the robot's help.  At level one, the user appears to need some help but the robot should not intervene yet so as to not interrupt the user.  This simplification addresses our primary interest -- detecting when the robot should help.  See the Future Work to see how we envision using all the levels of need.

\begin{table}[]
    \centering
    \begin{tabular}{c|c|c|c}
    Model           & Precision   & Recall    & F1    \\
    \hline
    Mutual Gaze         & .37   & .07       & .12   \\
    Confirmatory Gaze   & .65   & .04       & .08   \\
    Language            & .61   & .76       & .68   \\
    Multimodal Fusion   & .77   & .81       & .79   \\
    \end{tabular}
    \caption{Performance of each of the need detection models}
    \label{tab:results}
\end{table}

The gaze models do not perform well on their own, and their performance is noticeably worse than the previous gaze models \cite{kurylo2019using}.
Some of the decline in performance is expected.  When moving to automatically processed sensor data, instead of manual annotations, some degradations are expected.  There are also some differences between the current model and prior versions. The current model is based purely on eye gaze, whereas previous model combined user eye gaze with whether or not the robot was speaking.  Removing the robot speech diminishes the precision of the model, and future renditions will need to add this back into the model.

Overall, the language model performed relatively well and considerably better than the previous model, which had an F1 score of .08 \cite{reneau2020supporting}.  The addition of features to track question words and negations did not positively contribute to the results of this model.  Removing these features (one or both) showed a slight increase in the performance of the model.  

Even though the gaze models do not perform well on their own, they positively contribute to the fusion model.  The multimodal fusion is an improvement over the best performing model, the language model.  This shows that the addition of the gaze models improved the overall performance.

The performance of the fusion model is promising, but it is not sufficient to show that the system works in real-time.
To demonstrate that the trained system is capable of recognizing when a user needs assistance, we show a robot assisting a user in a different task.  In the video \footnote{https://youtu.be/eW2uVBgi9r4}, the robot is assisting a user who is preparing to bake cookies.  The user is at a step in the recipe that requires some vanilla to be added next.  The robot currently is not designed to recognize what step in the recipe the user is at (this is given to the robot), but it does use the system described here to recognize that the user needs help.  The provided video shows two instances of a user needing help, causing the robot to give some assistance.  The video also shows increasing amount of help provided by the robot, which is a feature that will be tied to the level of need in future versions of the architecture.

\section{Discussion}


For socially assistive robots to provide assistance that supports the fluid and natural progress of a task, is not disruptive, and supports the autonomy of the user, the robot must be able to decide when it is an appropriate time to intervene and provide some assistance.  
Using social cues from the user's eye gaze and language can provide a means of detecting when a user could use the help of the robot.
We present here a multimodal architecture that fuses together specialized models for mutual gaze, confirmatory gaze, and language to detect when a user needs help from the robot.
Results of the fusion model, trained on data collected in a Lego building task, indicate that the architecture is capable of detecting when a user needs help.  A demonstration of the architecture running with a social robot in a cooking scenario demonstrates the applicability of the architecture to some other tasks.

To train and evaluate our architecture, we conducted an experiment in which participants built a Lego structure while receiving assistance from a social robot.  Using this data, we first evaluated the specialized models.  The eye gaze models performed poorly (F1 = .12, F1 = .08), while the language model was considerably better (F1 = .68).  However, the fusion of the outputs of these models resulted in an even better model (F1 = 0.79).

These results are comparable with that of Reneau and Wilson \citeyear{reneau2020supporting}, who found that the fusion model produced an F1 up to .82.  Their result also improved over that of the gaze and language models, which all had F1 scores below .20.  However, our architecture and models differ from theirs in four ways:

\begin{enumerate}
    \item All of our input data for eye gaze and language is automatically processed, whereas Reneau \& Wilson relied on manual annotations of videos.  Our input data is likely less clean but more ecologically valid.
    \item In an effort to reduce our dependencies on the specifics of the task, our architecture does not include a model of the task.  For Reneau \& Wilson, a task model was able to detect when a user needs help with an F1 of .52.  This was a key factor in the performance of the fusion model.  Our fusion model does not integrate a task model, and yet our performance nearly matches theirs.
    \item A significant improvement over the prior work is the use of a learned language model.  Reneau \& Wilson employed a simple keyword spotting approach that resulted in an F1 of .08 on its own.  Our naive Bayes model performed relatively well on its own, achieving a F1=.68.  The performance of this model likely was a leading factor in making up for the lack of a task model in our architecture.
    \item Our gaze models have much lower precision than that of Reneau \& Wilson.  This is primarily due to our models not taking into account whether the robot is speaking.  Integration of the robot's speech would likely cause a significant improvement in these models.
\end{enumerate}

One important design choice in our architecture is the use of decision-level fusion instead of feature-level fusion, which would combine all the features into a single feature set and eliminate the specialized models for eye gaze and language models.  
However, the benefits of using specialized models include being able to integrate models specific to a modality, allowing the system to adapt to available data sources, and provide the potential for greater explanatory power.

Models that are specifically trained on a single modality (e.g., gaze or language) can leverage data where only that modality is available.  For example, our language model can be enhanced with additional training data, where the data might not have associated gaze data available.  Conversely, a feature-level fusion would require all data sources to be available.

We also envision the architecture to be able to adapt to available data streams without requiring a significant amount of retraining.  With a decision-level fusion, adapting the system to handle an additional specialized model will require retraining of only the fusion model. 

Decision-level fusion also provides the possibility of tracing the inference from the fusion model to the outputs of the specialized models, providing some justification for a model's inference.
In other words, when the robot provides help because the fusion model indicates that the user needs help, the inference that the user needs help can be traced to the inputs of the fusion.  If the user were to ask why the robot helped, the robot could inspect the results of the language and gaze models to generate a causal explanation of its behavior.
For example, a robot could specify that it concluded the user needed help because the user was exhibiting a confirmatory gaze behavior.
The explanations provide content on which a user can provide feedback to the robot, which could be integrated into the robot's future reasoning to improve inferences and adapt to user preferences \cite{wilson2020knowledge}.


One of the design goals of our architecture is to provide an approach that is applicable to many tasks in which a social robot is assisting on a physical task that does not require the robot to physically intervene.
Our evaluation shows that we can accurately recognize when a user needs help in a Lego building task, which should be analogous to other construction scenarios, like furniture assembly.
For cooking, we do not have a formal evaluation, but our demonstration is intended to show the applicability of our approach to cooking and  similar tasks.
Additionally, our architecture is based on previous work that used a medication sorting task \cite{reneau2020supporting}.
In applying our architecture to other tasks, we expect our gaze models to have a similar utility since patterns of gazing at a robot and a task would be present in other tasks.  We note that our gaze models made assumptions about the location of the robot and task, and we will need to  have automatic recognition of where those locations are for our system to easily transfer to a new task or environment.
Our language model has a risk of being task dependent since the content of the speech will have many details that are specific to the task.  In the short term, we plan on integrating language captured from other tasks, such as that used by Reneau \& Wilson \citeyear{reneau2020supporting}.  We also envision how integration with large, transformer models (i.e., GPT-3 \cite{brown2020language}) would minimize most task dependencies.

\subsection{Limitations}

A limitation of the current work is that the ground truth is the  analysis of a single annotator.  While labeling followed a well defined process, the subjectivity interpretation of the user's state warrants the use of additional annotators.  
Using the work of three annotators would allow us to employ a majority vote on all labels and increase the likelihood of accurately representing the state of the user.

\subsection{Future Work}

The current set of labels includes five levels of perceived user need.  However, our models reduce this to a binary classification problem of detecting when a user needs the robot's assistance.  In future work we seek to use all of the levels of need and be able to output a continuous value that is an estimate of the user's level of need.  A continuous value for need will allow the robot to choose an amount of assistance that is proportional to the amount of need \cite{wilson2019developing}.  thus enabling the robot to provide multiple levels of assistance \cite{Barnes2008, Greczek2015, wilson2016designing}.
 
Additional modalities are likely to improve the accuracy of the overall system. In \cite{reneau2020supporting}, they integrated the gaze and language models with a model that tracks the user's progress in the task.
While we currently are interested in an approach that is not specific to a task, we also plan on exploring extensions to the architecture that would allow for easy integration of task-specific models.

Emotions are also likely good indicators of when a person needs assistance.
If a user appears frustrated or angry, the user may be having difficulty with the task and needs assistance. 
Conversely, if the user appears engaged, interested, or intrigued, the user probably should not be interrupted with any assistance.  
To recognize the emotions, Grafsgaard et al. used facial expressions, posture, and hand gestures \cite{grafsgaard2014additive}.

\section{Conclusion}

In this paper, we present a real-time multimodal architecture to detect when a user needs assistance from a social robot. The approach taken in the architecture is to base the inference on the user's social cues and not be tied to specifics of the given task.  The design uses two eye gaze models, a language model, and a fusion model.  The system is trained and evaluated on data collected in an experiment in which a social robot assists with a Lego building task.  The results of the evaluation indicate that the architecture is capable of detecting when a user needs help. 
By recognizing a user's needs through these social cues, we provide a means for assistive agents to give more timely help, contributing to better support the user.

\bibliography{main}

\end{document}